# Application of Support Vector Machine Modeling and Graph Theory Metrics for Disease Classification

Jessica M. Rudd, MPH, GStat
Department of Statistics and Analytical Sciences
College of Science and Mathematics
Kennesaw State University
jrudd1@students.kennesaw.edu

*Abstract*— Disease classification is a crucial element of biomedical research. Recent studies have demonstrated that machine learning techniques, such as Support Vector Machine (SVM) modeling, produce similar or improved predictive capabilities in comparison to the traditional method of Logistic Regression. In addition, it has been found that social network metrics can provide useful predictive information for disease modeling. In this study, we combine simulated social network metrics with SVM to predict diabetes in a sample of data from the Behavioral Risk Factor Surveillance System. In this dataset, Logistic Regression outperformed SVM with ROC index of 81.8 and 81.7 for models with and without graph metrics, respectively. SVM with a polynomial kernel had ROC index of 72.9 and 75.6 for models with and without graph metrics, respectively. Although this did not perform as well as Logistic Regression, the results are consistent with previous studies utilizing SVM to classify diabetes.

*Keywords—support vector machine, logistic regression, graph theory, diabetes, disease classification*

## I. Introduction

Disease classification is a crucial element of biomedical research. Improved disease classification models aim to provide accurate and timely prediction to allow for earlier diagnosis and implementation of preventative measures. For example, in the US, approx. 29.1 million people (9.3%) are affected by diabetes, with 1/3 unaware of their disease status, and 57 million with pre-diabetes[1]. Diabetes and pre-diabetes are known to increase the risk of heart disease and stroke[1] but these long-term effects can be prevented with lifestyle changes and/or medical intervention[2]. Early screening and predictive risk models built with simple clinical measurements (no lab tests required) are important for deployment of prevention strategies, especially in undiagnosed population[3].

Traditionally, biomedical data is modeled using Logistic Regression, a method that relies on fitting data to a pre-determined model. Alternatively, the Support Vector Machine (SVM) algorithm is a supervised machine learning method that is a "model-free" method that does not require assumptions of distribution and interdependency of predictor variables. In SVM each data point is represented as a n-dimensional vector and the algorithm constructs an n-1-dimensional separating hyperplane to discriminate 2 classes, with maximized distance between the hyperplane and data points on each side. Non-linear functions, kernels, can also be used to transform data into multidimensional space. Previous research demonstrates that SVM has similar or improved predictive capabilities for disease classification in comparison to Logistic Regression[4].

In addition, it has been found that graph theory metrics provide useful information for the disease classification problem. Studies classifying diseases such as Alzheimer's[5] and Multiple Sclerosis[6] combined graph theory with machine learning methods, such as SVM, for improved prediction. I have not found previous research on the application of graph theory metrics to demographic and behavioral data for the prediction of disease.

This project aims to assess the application of SVM for classification of diabetes in a sample of people in Georgia, and apply graph theory metrics as potential predictors of disease in the model. This paper includes: Section II description of the dataset, Section III.A overview of SVM application to disease classification, Section III.B overview of graph theory application to disease classification, Section IV.A social network simulation, Section IV.B SVM algorithm, Section IV.C Logistic Regression algorithm, Section V results, and Section VI discussion.

## II. Dataset Details

Data from 2015 were obtained from Georgia's Behavioral Risk Factor Surveillance System (BRFSS)[7], landline and cellphone based survey conducted by the Centers for Disease Control and Prevention (CDC). A binary predictor variable was defined based on survey respondents reporting they had been informed by their physician they had diabetes or pre-diabetes. Once imputing missing data where possible, the analysis dataset included 2066 observations with 401 (19.4%) classified as having diabetes or pre-diabetes. BRFSS includes over 300 variables of various health behaviors and chronic conditions. For this study, I selected potential predictor variables based on literature review and the known conceptual model[8]. Variables considered based on the conceptual model included: sex, age, race, education, income, marital status,

BMI, cholesterol, hypertension, arthritis, physical activity, and consumption of fruits and vegetables. Once cleaning the predictors of interest, the analysis dataset included a sample of 1284 people and households in Georgia.

III. LITERATURE REVIEW

*A. SVM application to disease classification*

Yu, et al.[4] used the National Health and Nutrition Examination Survey (NHANES), an ongoing, cross-sectional, probability sample of US population, to build SVM and Logistic Regression classification models for 2 classification schemes: persons with diabetes (diagnosed or undiagnosed) vs. persons without diabetes, and persons with undiagnosed diabetes or pre-diabetes vs. persons without diabetes. They used 14 potential predictors commonly associated with diabetes: family history, age, gender, race and ethnicity, weight, height, waist circumference, BMI, hypertension, physical activity, smoking, alcohol use, education, household income[4]. They found that the Radial Basis Function (RBF) kernel, and Linear kernel worked best for classification schemes I and II respectively, and there was no significant difference between Logistic Regression and SVM performance (AUC 0.83 & 0.73 for classification I and II, respectively, with both models)[4].

Additionally, Kumari et al[9]. also found success with the SVM model for classification of diabetes in the Pima India Diabetic Dataset from the UCI Machine Learning Laboratory. In this case, an 8 predictor SVM model, including lab data (plasma glucose concentration, 2-hr serum insulin) was validated with 78% accuracy using the RBF kernel.

SVM has been used across diverse biomedical classification problems. This includes a patient financial risk model using health claims and clinical encounter data, and a patient response to flu awareness campaign model, both using weighted SVM[10]. A project comparing various machine learning techniques with Logistic Regression for prediction of heart disease also shows no significant difference between Logistic Regression and SVM, with the Linear kernel performing best[11].

*B. Graph Theory application to disease classification*

In Kocevar, et al. [6]. they combined graph metrics with SVM to classify various Multiple Sclerosis (MS) clinical profiles. Cortical and sub-cortical gray matter (GM) segmentation was performed on the advanced MRI imaging of 77 MS patients and 26 healthy controls (HC). Figure 1 shows the process of segmentation of the scans to create nodes, and anatomically constrained probabilistic streamline tractography is used to create edges between the segments. Edge weights are determined by a function of the number of fibers connecting the segments. The weakest connections are removed by applying a threshold $0 \leq \tau \leq 1$ on the weighted graph, generating an unweighted graph maintaining the τ% strongest connections in the network. Global network metrics calculated for the study on the unweighted graph included: graph density ($D$), assortativity ($r$), transitivity ($T$), global efficiency ($E_g$), modularity ($Q$), and characteristic path length (CPL).

FIGURE 1. BRAIN IMAGING GRAPH GENERATION

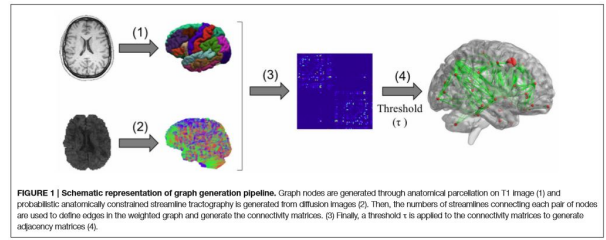

The study found that global graph metrics were not significantly dependent on patients' age or gender. Overall, significant difference in graph metrics were found when comparing MS patients with HC groups, as well as between different clinical classifications of MS. SVM classification with RBF kernel was then used to predict varying binary classifications of HC groups and clinical courses, with highest classification achieved using all graph metrics as a feature vector in the model at 91.8%. Using only one graph metric, the best in this case being modularity, the study could achieve accuracy of 88.9%.

In Khazaee, et al., they found that using changes in brain connections from functional magnetic resonance imaging (MRI) provided strong predictive measures for classifying Alzheimer's Disease (AD) patients from healthy controls (HC). 20 patients with AD and 20 age-matched HC from Alzheimer's disease neuroimaging initiative (ADNI) database were selected for study. MRI images were parcellated into 90 regions and edges were defined as connectivity of all pairs of regions using Pearson's correlation coefficient. As in the previous study, thresholding was used to maintain the strongest connections in the network. Preserving a high proportion of the network results in a dense graph with noisy and less significant edges maintained. However, removing too many edges can result in a disconnected graph where global graph metrics cannot be calculated. From their previous research, this study found that a threshold of 12% was optimal[5]. The study maintained the bridge edges between any disconnected sections that resulted from thresholding, regardless of the edge weight. Figure 2 shows the weighted adjacency matrix for the complete and 12% threshold network.

FIGURE 2. WEIGHTED ADJACENCY NETWORKS

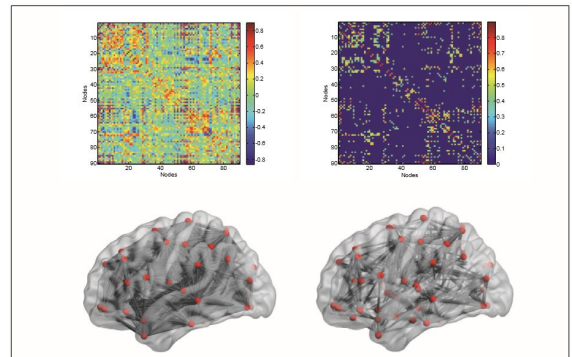

Figure 1. Topology structure of functional connectivity network of one healthy subject. (Left): The connection matrix (top) and topology (bottom) of the fully connected network. (Right): Same as (Left) except only 12% of edges is retained. Plots of this figure were created using the BrainNet Viewer software package (http://nitrc.org/projects/bnv/).

Graph metrics calculated for this project included: functional segregation via clustering coefficient, local efficiency, and normalized local efficiency to measure specialized processing within densely interconnected groups of regions; functional integration via characteristic path length and global efficiency to assess ability of the brain to rapidly combine specialized information from distributed regions; and 3 local measures including degree, participation coefficient, and betweenness centrality to measure properties of the 90 regions. An iterative feature selection algorithm using 7 different methods was then used to filter the most effective graph features for the classification problem. Linear SVM with a tuned C parameter using leave-one-out cross validation was used to perform the final feature classification.

End results found that Fisher Score provided the best feature selection method for the discriminative algorithm. Figure 3 shows performance of the Fisher algorithm with increasing number of selected features and various values of SVM C parameter. The best algorithm found could classify AD patients from HC group with a highest accuracy of 97.5%.

FIGURE 3. PREDICTIVE PERFORMANCE OF VARIOUS # OF PREDICTIVE FEATURES AND C PARAMETER SELECTION

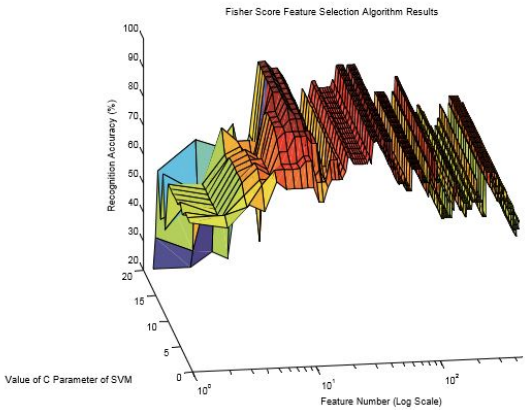

## IV. METHODOLOGY

SAS® PROC HPSVM was used to build SVM models. PROC HPSVM is a SAS® Enterprise Miner™ high performance data mining procedure built to take advantage of parallel processing with both single machine and distributed multiple-machine mode. The data were split into 80% training, 20% validation datasets. I ran PROC SVM comparing 3 common kernels: Linear, Polynomial, and RBF. I used 5-fold cross validation for each kernel to determine the best penalty parameter, C. This controls for overfitting of the model by specifying allowable misclassification. SAS® PROC LOGISTIC was used to build the Logistic Regression model for comparison. The models were compared based on sensitivity, specificity, and area under the receiver operating characteristic (ROC) curve, using Enterprise Miner™.

### A. Social Network Simulation

To test the application of graph theory metrics as potential predictors in a classification model, it was necessary to simulate a social network within the BRFSS dataset. The Watts-Strogatz small world network model was selected to represent the sample social network for this application due to its ability to simulate the interconnected groups (clusters) that exist in real-world networks, as well as the existence of random irregular connectivity patterns[12]. This model was first introduced by Duncan Watts and Steven Strogratz in *Nature* in 1998[13].

Watts-Strogatz is a variation of the lattice network where nodes are connected to their nearest neighbors only. Figure 4 illustrates the adjustment of a lattice network to the Watts-Strogatz network. Watts-Strogatz randomly rewires some of the lattice edges, resulting in high clustering and short paths. This network is undirected. Ideally, the data studied would include some network characteristics, but I did not have access to a public use dataset that includes both demographic and network characteristics. For the purposes of testing the application of graph metrics to a predictive model, a simulated network will suffice. To incorporate an element of the demographic data into the social network simulation, I weighted each edge by the average standardized number of adults in the respondents' household.

FIGURE 4. LATTICE AND WATTS-STROGATZ NETWORK MODELS

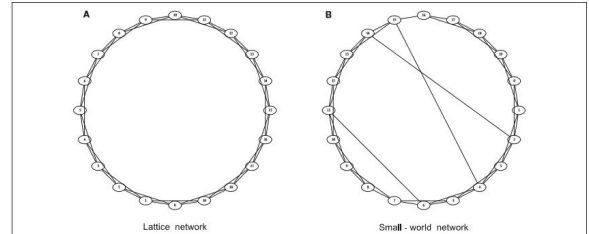[12]

The algorithm for creating a Watts-Strogatz network starts with a lattice network where each node is adjacent to a defined $L$ neighbors. If each node has degree $k_L$, and $k_L$ is even, then the global clustering coefficient, $\overline{C}$, of the network is

$$\overline{C} = 0.75(k_L - 2)/(k_L - 1) \qquad (1)$$

To randomly rewire the lattice network, each edge has a defined probability, $p_w$, of being re-wired. Each edge can only have one end re-wired and the edges are replaced so that total number of edges and mean degree is the same as the original lattice[12]. The reason why the Watts-Strogatz model maintains high clustering coefficient, but low average path length, as compared with a random network, is that the global clustering coefficient is based on the average of the local measure; rewiring a small number of connections will only affect the local clustering coefficient of a small number of nodes. However, average path length is a global measure of the average of shortest path length between every combination of nodes. Changing even one edge can create shortcuts between many pairs of nodes, greatly affecting the average path length[12]. In addition, even if we select a rewiring probability of 1, a Watt-Strogatz network will not be the same as a random network with same size and average degree because the Watts-Strogatz algorithm does not allow nodes to have degree less than n/2, where the random network does allow this[12].

Figure 5 shows the simulated BRFSS network for 10% of the data.

FIGURE 5. WATTS-STROGATZ SIMULATED NETWORK

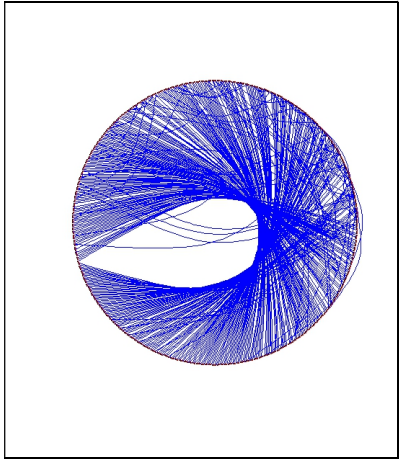

The R igraph package was used to create the Watts-Strogatz network using 1284 nodes, $k_L = 4$ (degree of every node in the initial lattice), and rewiring probability $p_w$ 0.5. The resulting edge-list was imported into SAS® and merged with the household variable in the analysis dataset to create the edge weights as described above. SAS® PROC OPTGRAPH was used to calculate various graph characteristics to be used in the SVM and Logistic Regression model application, including: local clustering coefficient, degree centrality, closeness centrality, betweenness centrality, and eigenvector centrality.

*B. Support Vector Machine*

SVM is a supervised learning algorithm that represents instances of data as points in space and then builds a model to assign new instances to one category or another. Each data point is represented as a n-dimensional vector, then SVM constructs an n-1-dimensional separating hyperplane to discriminate 2 classes, with maximized distance between the hyperplane and data points on each side. SVM aims to find the best hyperplane for separation of both classes[11].

Data are represented as

$$(\vec{x}_1, y_1), \ldots, (\vec{x}_n, y_n) \quad (2)$$

where $y_i$ is either 1 or -1, indicating to which class $x_i$ belongs. Each $x_i$ is $p$-dimensional vector representing all of the characteristic values (variables) of $x_i$. The hyperplane that best separates the group of $x_i$ vectors where $y_i = 1$ from the group of vectors where $y_i = -1$ is

$$\vec{w} \cdot \vec{x} - b = 0, \quad (3)$$

Where $\vec{w}$ is the normal vector to the hyperplane and $b$ is the offset of the hyperplane from the origin. If the data points are linearly separable, the hard margin can be represented as

$$\vec{w} \cdot \vec{x} - b = 1$$

and

$$\vec{w} \cdot \vec{x} - b = -1. \quad (4)$$

Figure 6 shows a maximum margin separation for linearly separable data. The samples that fall on the margin are known as the support vectors.

FIGURE 6. MAXIMUM MARGIN HYPERPLANE

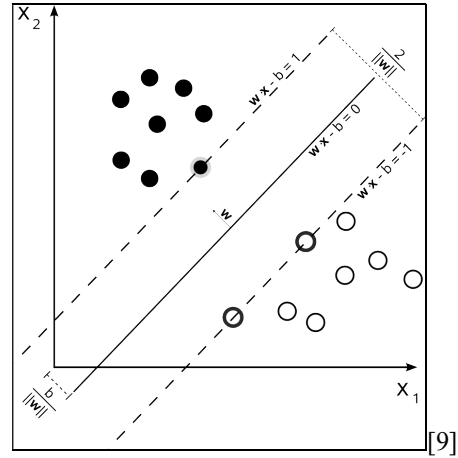
[9]

For data that is not linearly separable we can include a hinge loss function, 'C', to determine the trade-off between increasing the margin and whether an instance of $x_i$ lies on the correct side of the margin. In addition, we can implement kernel functions to adjust the inner dot product of the maximum margin hyperplane optimization algorithm. This transforms the data into a higher dimensional space. Figure 7 shows the transformation of features into higher dimension space.

FIGURE 7. KERNEL TRANSFORMATION OF FEATURE SPACE

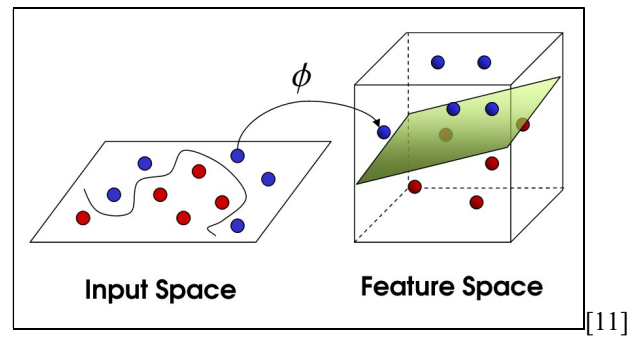
[11]

SAS® PROC HPSVM from the Enterprise Miner™ High Performance Procedures was used to build the SVM models. 5-fold cross validation was used to determine best soft margin parameter C in each kernel: Linear, RBF, and Polynomial. Macro programming was used to automatically evaluate several values of RBF kernel parameter, γ, and then choose the best C/γ combination.

## C. Logistic Regression

Logistic Regression examines the non-linear relationship between a binary outcome and categorical or continuous predictor variables. The logistic model outputs a probability of an event between 0 and 1 as the log of the odds ratio (3)

$$Ln\left(\frac{P}{1-P}\right) = \beta_0 + \beta_1 X_1 + \beta_2 X_2 + \ldots + \beta_k X_k \quad (5)$$

where β is the parameter coefficient and x is the value of the independent variable.

SAS® PROC LOGISTIC was used to build the model and stepwise elimination with α = 0.05 was used to eliminate redundancy and keep the strongest predictors in the model. 10-fold cross validation was used for model evaluation.

## V. RESULTS

Tables 1 and 2 illustrate the significant effects remaining in the Logistic Regression models, for models with and without network characteristic metrics included. Age, BMI, hypertension, and cholesterol all have increased odds of diabetes outcome, while education has decreased odds. This is consistent with outcomes of previous research and known risk factors for diabetes. In the model with graph metrics included, only closeness centrality remains as significant, in addition to the same demographic variables from the previous model. If this were a real network in the dataset (not simulated), this would indicate that people with shorter total paths to other people in the network would have increased risk of diabetes.

TABLE 1. LOGISTIC REGRESSION SIGNIFICANT EFFECTS

| No network characteristics | | |
|---|---|---|
| Odds Ratio Estimates and Wald Confidence Intervals | | |
| Effect | Estimate | 95% CI |
| Age | 1.62 | (1.26, 2.09) |
| Education | 0.58 | (0.47, 0.71) |
| BMI | 3.77 | (1.96, 7.22) |
| Hypertension | 3.32 | (2.08, 5.30) |
| Cholesterol | 3.48 | (2.24, 5.40) |

TABLE 2. LOGISTIC REGRESSION SIGNIFICANT EFFECTS

| Including network characteristics | | |
|---|---|---|
| Odds Ratio Estimates and Wald Confidence Intervals | | |
| Effect | Estimate | 95% CI |
| Closeness centrality | 1.39 | (1.11, 1.73) |
| Age | 1.55 | (1.20, 1.99) |
| Education | 0.57 | (0.46, 0.68) |
| BMI | 3.74 | (1.95, 7.19) |
| Hypertension | 3.40 | (2.12, 5.44) |
| Cholesterol | 3.37 | (2.16, 5.24) |

Table 3 shows the model performance results for models with and without the social network graph characteristics. The models were evaluated based on the sensitivity, specificity, and ROC index for the validation data set. The logistic model performs best for models with and without graph metrics included, and the SVM model with polynomial kernel is comparable but the ROC index is affected by lower sensitivity. Figures 8 and 9 provide a visual comparison of the area under the ROC curves for models with and without graph metrics, respectively.

TABLE 3. LOGISTIC REGRESSION AND SVM RESULTS

| MODEL | GRAPH METRICS | KERNEL | BEST C | SENSITIVITY | SPECIFICITY | ROC INDEX (TEST) |
|---|---|---|---|---|---|---|
| LOGISTIC | NO | NA | NA | 34.2 | 97.7 | 81.7 |
| LOGISTIC | YES | NA | NA | 36.8 | 97.2 | 81.8 |
| SVM | NO | LINEAR | 0.2 | 0 | 100 | 80.4 |
| SVM | YES | LINEAR | 0.5 | 0 | 100 | 78.9 |
| SVM | NO | POLY | 0.2 | 26.3 | 96.4 | 75.6 |
| SVM | YES | POLY | 0.2 | 23.7 | 97.3 | 72.9 |
| SVM | NO | RBF | 1 | 0 | 100 | 70.9 |
| SVM | YES | RBF | 0.2 | 0 | 100 | 71.1 |

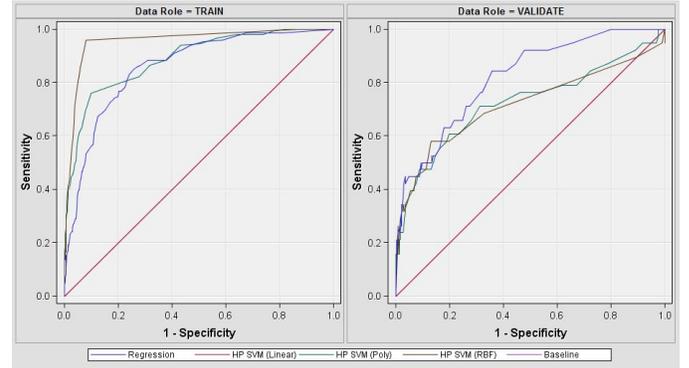

FIGURE 8. COMPARISON OF ROC CURVES: MODELS INCLUDING GRAPH METRICS

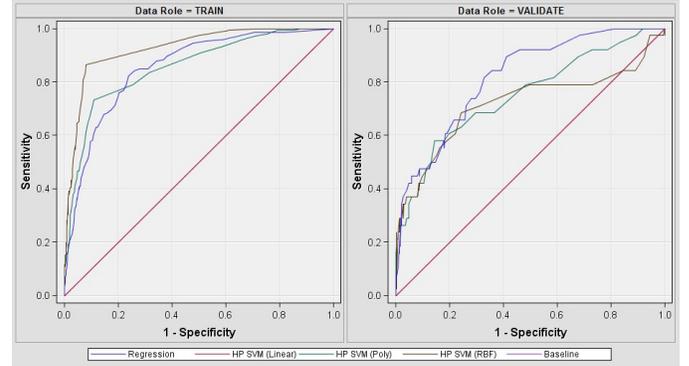

FIGURE 9. COMPARISON OF ROC CURVES: MODELS NOT INCLUDING GRAPH METRICS

## VI. DISCUSSION & FUTURE RESEARCH

Support Vector Machines and graph metrics are important tools to be considered for disease classification problems. While SVM did not perform as well as Logistic Regression in this study, it's results were comparable to previous research. SVM is known to be less sensitive to high dimensionality, and sparse datasets, so would likely perform better than Logistic Regression in studies with biomedical data of that nature.

Including graph metrics in the model did improve predictive performance slightly using a simulated network. Ideally, future research will include a dataset with both demographic and network characteristics included. Future improvements to this study will include:

- Parameter selection using machine learning such as Random Forest
- Creating a custom kernel for SVM based on conceptual model of diabetes
- Creating a custom kernel for SVM using deep learning techniques
- Performing grid search for improved C and gamma optimization for SVM kernels

## VII. RELEVENT SAS® CODE

### A. R code for Watts-Strogatz network generation

```
library(igraph)
network<-watts.strogatz.game(1, 1284, 5, .5, loops = FALSE,
                              multiple = FALSE)
edges <- as_edgelist(network)
colnames(edges) <- c("ID_1", "ID_2")
write.csv(edges, file = "C:/users/Jess/OneDrive/Grad School/Graph Theory/Project/edges.csv",
          row.names = TRUE)
```

### B. SAS® PROC OPTGRAPH

```
/*Run node characteristics */
proc optgraph
    loglevel=moderate
    data_links = edges4
    out_nodes  = NodeSetOut;
    performance
       nthreads = 2;
    centrality
       clustering_coef
       degree    = OUT
       influence = weight
       close     = weight
       between   = weight
       eigen     = weight;
run;

%put &_OPTGRAPH_;
%put &_OPTGRAPH_CENTRALITY_;
```

### C. SAS® PROC HPSVM

```
%let predicts = HLTHPLN1 MARITAL SEX _DRDXAR1 _EDUCAG _INCOMG _PAINDX1 _RFBMI5 _RFCHOL
_RFHYPE5 _RFSMOK3 race _age_g;
/* SVM with polynomial kernel, various C */
proc hpsvm data=model method=activeset;
input &predicts /level=nominal order=asc;
input _FRUTSUM _VEGESUM/level=interval;
partition  fraction(validate=.2 seed=12345);
KERNEL POLYNOM / DEGREE=2;
penalty c= .2 .5 1 1.5 2;
select cv=random fold=5 seed=34567;
target diabetes;
output outclass=bigdata.outclass outfit=bigdata.outfit outest=bigdata.outest;
run;

proc svmscore data=model out=score
inclass=outclass infit=outfit inest=outest;
run;
```